\documentclass[conference]{IEEEtran}
\IEEEoverridecommandlockouts
\usepackage{cite}
\usepackage{amsmath,amssymb,amsfonts}
\usepackage{algorithm}
\usepackage{algpseudocode}
\usepackage{graphicx}
\usepackage{subfigure}
\usepackage{textcomp}
\usepackage{graphicx}
\usepackage{placeins}
\usepackage{multirow}
\usepackage{tabularx}
\usepackage{float}
\usepackage{array}
\usepackage{caption} 
\usepackage{xcolor}
\DeclareMathOperator*{\argmin}{argmin}
\def\BibTeX{{\rm B\kern-.05em{\sc i\kern-.025em b}\kern-.08em
    T\kern-.1667em\lower.7ex\hbox{E}\kern-.125emX}}

\expandafter\def\expandafter\normalsize\expandafter{%
    \normalsize%
    \setlength\abovedisplayskip{0pt}%
    \setlength\belowdisplayskip{8pt}%
    \setlength\abovedisplayshortskip{-8pt}%
    \setlength\belowdisplayshortskip{2pt}%
}

\begin{document}

\title{Online Learning of Multiple Tasks
and Their Relationships : Testing on Spam Email Data and EEG Signals Recorded in Construction Fields\\}

\author{\IEEEauthorblockN{Yixin Jin*}
\IEEEauthorblockA{
\textit{University of Michigan, Ann Arbor}\\
Ann Arbor, MI, 48109, USA \\
jinyixin@umich.edu}
\and
\IEEEauthorblockN{Wenjing Zhou}
\IEEEauthorblockA{
\textit{University of Michigan, Ann Arbor}\\
Ann Arbor, MI, 48109, USA \\
wenjzh@umich.edu}
\and
\IEEEauthorblockN{Meiqi Wang}
\IEEEauthorblockA{
\textit{Brandeis University}\\
Waltham, MA, 02453, USA \\
meiqw@brandeis.edu}
\and
\IEEEauthorblockN{Meng Li}
\IEEEauthorblockA{
\textit{Columbia University}\\
New York, NY, 10027, USA \\
ml4818@columbia.edu}
\and
\IEEEauthorblockN{Xintao Li}
\IEEEauthorblockA{
\textit{University of Miami}\\
Miami, FL, 33156, USA \\
xintao.li@miami.edu}
\and
\IEEEauthorblockN{Tianyu Hu}
\IEEEauthorblockA{
\textit{Columbia University}\\
New York, NY, 10027, USA \\
th3011@columbia.edu}
}

\maketitle

\vspace{-0.2cm}
\begin{abstract}
This paper examines an online multi-task learning (OMTL) method, which processes data sequentially to predict labels across related tasks. The framework learns task weights and their relatedness concurrently. Unlike previous models that assumed static task relatedness, our approach treats tasks as initially independent, updating their relatedness iteratively using newly calculated weight vectors. We introduced three rules to update the task relatedness matrix: OMTLCOV, OMTLLOG, and OMTLVON, and compared them against a conventional method (CMTL) that uses a fixed relatedness value. Performance evaluations on three datasets—a spam dataset and two EEG datasets from construction workers under varying conditions—demonstrated that our OMTL methods outperform CMTL, improving accuracy by 1\% to 3\% on EEG data, and maintaining low error rates around 12\% on the spam dataset.
\end{abstract}

\begin{IEEEkeywords}
machine learning, online learning, data mining, EGG
\end{IEEEkeywords}

\vspace{-0.2cm}
\section{Introduction}
\vspace{-0.18cm}

In the big data era \cite{zhu2021taming, gao2018solution, bu2021gaia, peng2023gaia}, the need for prompt data \cite{xin2024mmap} processing and decision-making has grown, particularly in areas like multi-task learning \cite{liu2024adaptive}. This paper addresses the shortcomings of traditional online learning models, which assume static task relatedness, by introducing a dynamic framework for Online Multi-Task Learning (OMTL). Our approach iteratively updates task relatedness based on data insights, improving the accuracy and utility of online learning systems in applications like spam detection and EEG signal analysis. Our research advances the understanding of task relatedness in OMTL and demonstrates its practical applications, highlighting its significance for real-time data processing in various settings \cite{read2023prediction}, \cite{read2022prediction}, \cite{shen2024localization}, \cite{li2024exploring}, \cite{feng2022beyond}, \cite{chen2022visual}, \cite{bu2016attention}. This study showcases the adaptability and impact of multi-task learning models in data-intensive environments.

\section{Materials and Methods}

\subsection{Datasets}
\subsubsection{Spam emails dataset}
To test the suggested framework in this research, the spam dataset provided online was used \cite{crammer2009adaptive} to classify spam and non-spam emails. This dataset includes 3,000 training and 1,100 testing examples recorded from email messages of two subjects. It includes approximately 50\% spam email messages and 50\% non-spam email messages. The first 4,000 training and testing examples are subject 1 data. The last 100 data are from the subject 2. Subject 1 and 2 were considered as task 1 and 2 respectively.

\subsubsection{Construction workers' EEG signal dataset}

The EEG data were collected from 8 healthy male workers using the Emotiv EPOC+, an affordable EEG device that records signals through 14 channels at a quality suitable for research. The device sampled internally at 2,048Hz, delivering data at 128Hz with a resolution of 14 bits and a connectivity at the 2.4GHz band. Two datasets were generated based on the motor cortex activation: Dataset1 distinguished between inactive (inactive) and active (active) states, while Dataset2 differentiated between relaxed and stressful working conditions. Details on window size selection and feature extraction will be discussed in the subsequent section.

\subsection{Dataset pre-process and feature extraction for EEG signal data}
\subsubsection{Pre-process}
As described in section A 2), the construction workers’ EEG signal dataset was raw data recorded as time-series data in 14 channels and was originally stored as an Excel file. A step of preprocessing is applied to extract the data into a Matlab format. Due to the raw and uncleaned natural of the original data, there is a large number of signal artifacts and abnormality in the extracted dataset. These signal artifacts were removed before extracting the features using filtering methods and the Independent Component Analysis method (ICA). After cleaning the data \cite{yang2024comparative}, a window of size 128 was applied to calculate the feature vectors. The window size was 128 since we are using a 128Hz rate to collect the data. Moreover, in each of the subject datasets, worker’s behavior is labeled up to 7 labels. We extracted two interested datasets that each has binary labels. Dataset1 contains the label indicating whether the constructor is rest or active. Dataset2 labels the constructor as stressful or relaxed. We used Python to handle all the extraction and cleaning and put the well-organized data into a .mat file just like spam data in order to process it. We have 2,744 data points in dataset1 and 1,585 data points in dataset2.

\subsubsection{Power spectral density estimation}
The power spectral density (PSD) estimation shows the strength of energy variation as a function of frequency. PSD is the average power distribution of frequency response of a random periodic signal. The power distribution of frequency is calculated through the following equations:
\begin{align}
S(w) =& \sum p(k)e^{iwk} \label{eq1} \\
\text{where } p(k) =& \sum y(t)y*(t-k) 
\end{align}
After calculating the power distribution, the PSD is calculated through averaging the absolute mean value over the frequency domain \cite{xin2023self}. In this study, we used $pwelch$ function in Matlab to calculate the power distribution and then apply the absolute mean to calculate the PSD. We apply the same PSD extraction function to generate a feature vector from multi-channel signals (14 channels in this paper) from different EEG electrodes (14 electrodes). This process is the same for the following feature extraction process, and therefore will not be repeated in the following sections.

\subsubsection{Mean estimation for alpha frequency and beta
frequency}
Alpha frequency and beta frequency are two sets of frequency domains with which to describe human brain activities. Alpha frequency describes the frequency between 8 to 12.5HZ, whereas beta frequency describes 12.5 to 30HZ. Based on research, alpha frequency is associated with the movement of closing eyes and beta frequency mainly describes the muscle contractions before and during a human movement; therefore, these are the best features with which to determine whether a human is in resting or active state.

To obtain alpha and beta frequencies for a specific channel of brain signals, we can apply the same power distribution function with the cut-off frequency domain setup to the corresponding frequency domains and apply the similar mean absolute estimate function to extract the mean estimation. 

\subsubsection{Frontal EEG features}
Frontal EEG asymmetry (FEA) compares the frontal activity of the brain between its left and right areas. Left frontal activity indicates a positive emotion; on the other hand, right frontal activity usually indicates a negative emotion \cite{winkler2010frontal}. FEA shows the degree of activation of the left and right areas by comparing the spectral power in the alpha and beta range frequencies between these two areas \cite{coan2006capability}. FEA has been frequently used to determine the emotions of human subjects to measure the valence and arousal of human subjects’ emotional levels \cite{allen2015frontal}. Valence levels illustrate how pleasurable the event is for the subjects, and arousal shows how active/aroused subjects are in different situations \cite{scherer2005emotions}. In order to calculate the arousal and valence features, we use the following, equation \ref{eq2} and equation \ref{eq3}: 
\begin{align}
Arousal =& \frac{\alpha (AF3 + AF4 + F3 + F4)}{\beta (AF3 + AF4 + F3 + F4)} \label{eq2} \\
Valence =& \frac{\alpha (F4)}{\beta {F4}} + \frac{\alpha (F3)}{\beta (F3)} \label{eq3} 
\end{align}
In both equations, $\alpha$ and $\beta$ represent the alpha and beta frequencies following the same calculation procedure in section B 3). The AF3, AF4, F3 and F4 represent the different EEG channels in the left and right brains. Equation \ref{eq2}, which is the arousal feature, indicates the excitation state of a person by calculating the beta/alpha ratio. Equation \ref{eq3}, the valence feature, compares the activation levels of two cortical hemispheres of the brain that shows the emotional valence status of the subjects.

\subsection{Perceptron based online multi-task learning (CMTL)}
In machine learning, the perceptron is an algorithm for supervised learning of binary classifiers that decide whether an input belongs to some specific class or not \cite{xin2024vmt}. The perceptron uses hypotheses of the form
$y(x;w) = f(w^T x)$, where $f(z)=I[z \geqslant 0]$. The update rule is as follows:
\begin{align}
w_{i+1} := w_{i} + \alpha[t_{i+1} - y(x_{i+1}; w_{i})]x_{i+1}
\end{align}
CMTL keeps a weight vector for each task and updates all weight vectors at each mistake using the perceptron rule through learning rates defined by a $K \times K$ interaction matrix A. It is A that encodes beliefs about the learning tasks: different choices regarding the interaction matrix result in different geometrical assumptions regarding the tasks. The pseudocode for the multitask perceptron algorithm using a generic interaction matrix A is given below. At the beginning of each time step, the counter $s$ stores the mistakes made so far, plus one. The weights of the K perceptrons are maintained in a compound vector $w^T_s = (w^T_{1, s}, ..., w^T_{K, s})$ with $w_{j, s} \in \mathbb{R}^d$ for all j. The algorithm predicts $y_t$ through the sign $y_t$ of the ith perceptron’s margin $w^T_{s-1} \Phi_t = w^T_{i, s-1} x_t$. Then, if the prediction and the true label disagree, the compound vector update rule is $w_s = w_{s-1} + (A \otimes I_d)^{-1} \Phi_t$, where $\otimes$ was defined as the $Kd \times Kd$ Kronecker product, that is $A \otimes I_d = 
\begin{bmatrix}
    a_{11}I_d      & a_{12}I_d & \dots & a_{1K}I_d \\
    \hdotsfor{4} \\
    a_{K1}I_d      & a_{K2}I_d & \dots & a_{KK}I_d
\end{bmatrix}$ Since $(A \otimes I_d)^{-1} = A^{-1} \otimes I_d$, the above update is equivalent to the K task updates \cite{dekel2006online}:
\begin{align}
w_{j,s} = w_{j, s-1} + y_t A^{-1}_{j,i_t}x_t
\end{align}

The pseudocode is shown as algorithm \ref{alg:cmtl}.

\begin{algorithm}
\small
\caption{Perceptron based online multi-task learning (CMTL)}\label{alg:cmtl}
\begin{algorithmic}[1]
\State \textbf{Parameters:} Positive definite $K \times K$ interaction matrix A.
\State \textbf{Initialize:} $w_0 = 0 \in \mathbb{R}^{Kd}, s=1$
\For{$t = 1, 2,3...$}
\State Observe task number $i_t \in \{1,...,K\}$ and the instance vector $x_t \in \mathbb{R}^{Kd}: ||x_t|| = 1$
\State Build the associated multitask instance $\Phi_t \in \mathbb{R}^{Kd}$
\State Predict label $y_t \in \{-1, +1\}$ $y_t \in \{-1, +1\}$ with $\hat{y}_t = \text{SGN} (w^T_{s-1} \Phi_t)$
\State Get label $y_t \in \{-1, +1\}$ 
\If{$\hat{y}_t \neq y_t$}
\State Update $w_s = w_{s-1} + y_t (A \otimes I_d)^{-1} \Phi_t$
\EndIf
\EndFor
\end{algorithmic}
\end{algorithm}

\vspace{-0.2cm}
\subsection{Online multi-task learning (OMTL)}
\vspace{-0.2cm}
This method learns weight vectors of multiple tasks and the task relatedness matrix together adaptively in an online setting, in contrast with previous methods, which usually assume the task relatedness matrix is fixed and the relatedness of task is positive. First, this method defines an objective function to optimize, which is inspired by the CMTL method. The objective function is as follows:
\vspace{0cm}
\begin{align}
\argmin_{w \in \mathbb{R}^{Kd}} \frac{1}{2}w^T A_\otimes w + D_A(A||A_t) + \sum^{t}_1 l_t(w)
\end{align}
\vspace{0cm}

where $D_A$ denotes Bregman divergences, w and A are the weight vector and the interaction matrix, and $A_\otimes = A \otimes I_d$ 
The optimization problem is defined jointly over both w and A. It can be solved in an alternating fashion by solving for w given A, and then solving for A given w \cite{saha2011online}. So this method uses an alternating optimization scheme to solve for the parameters A and w. Deriving from the CMTL method, the update rule is as follows:
\vspace{0cm}
\begin{align}
w_s &= w_{s-1} + y_t (A_{s-1} \otimes I_d)^{-1} \Phi_t \\
w_{j,s} &= w_{j, s-1} + y_t A^{-1}_{s-1,(j,i_t)}x_t 
\end{align}
\vspace{0cm}

where j denotes which task the parameter is for, s standing for the round, true label $y_t \in \{-1, 1\}$, $\Phi_t = (0,...,0,x_{i_t},0,...,0) \in \mathbb{R}^{Kd}$.

Once $w_s$ is solved, we treat it as fixed and then solve for A. The pseudocode for OMTL is shown in algorithm \ref{alg:omtl}.

\begin{algorithm}
\small
\caption{Online multi-task learning (OMTL)}\label{alg:omtl}
\begin{algorithmic}[1]
\State \textbf{Input:} Examples from $K$ tasks, Number of rounds.
\State \textbf{Output:} $w$ and a positive definite $K\times K$ matrix $A$, learned after $T$ rounds.
\State \textbf{Initialization:} $A = \frac{1}{K} \times I_d; w_0 = 0;$
\For{$t = 1, ..., T$}
\State receive the pair $(x_t, i_t), x_t\in \mathbb{R}^{d}$
\State construct $\Phi_t \in \mathbb{R}^{Kd}$ from $x_t$
\State predict label $\hat{y}_t = SGN(w^T_{s-1} \Phi_t) \in \{-1, +1\}$
\State receive true label $y_t\in \{-1, +1\}$
\If{$\hat{y_t} \neq y_t$}
\For{$j=1,...,K$}
\State $w_{j,s} = w_{j, s-1} + y_t A^{-1}_{s-1,(j,i_t)}x_t$
\EndFor
\If{$t>$EPOCH}
\State update $A_s$
\State $s \gets s + 1$
\EndIf
\EndIf
\EndFor
\end{algorithmic}
\end{algorithm}

\vspace{-0.2cm}
\section{Results \& Discussion}
\vspace{-0.2cm}
We evaluated the developed online task relationship learning algorithm by comparing five different algorithms suggested in this paper \cite{saha2011online}; these algorithms have been shown in the Table \ref{eval}. We tested our algorithm on spam and two EGG dataset. The evaluation metric we used is the error rate of classification, which is the total number of correct classifications divided by the total number of samples. To fully investigate our algorithms, we first run our algorithms with fixed epoch parameters, which we made to 0.5, 0.8 and 0.8 respectively, and tuned the learning rate $\eta$ simultaneously. Also, we tried to explore the influence of the epoch parameters measuring the proportion of data being read before updating relatedness matrix. Therefore, we run our algorithm using various values for epoch parameters to investigate this aspect.

\begin{table}[H]
\caption{Description of the algorithm used to evaluate the performance of the developed online learning algorithm.}
\label{eval}
\begin{center}
\scalebox{0.76}{
\begin{tabular}{|l|l|}
\hline
\multicolumn{1}{|c|}{\bf Methods}  & \multicolumn{1}{c|}{\bf Description} \\
\hline
CMTL &Online perceptron with fixed interaction matrix \\
\hline
BatchOPT & Online multitask perceptron with batch optimal update for matrix A\\
& $A_s = \frac{(W^T_{s - 1}W_{s-1})^\frac{1}{2}}{tr((W^T_{s - 1}W_{s-1})^\frac{1}{2})}$ \\
\hline
OMTLCOV             &Online multitask perceptron with covariance based update for matrix A\\
& $A_s = cov(W_{s-1})$\\
\hline
OMTLLOG             &Online multitask perceptron with LogDet divergence based update for matrix A\\
& $A_s = (A^{-1}_{s-1} + \eta sym(W^T_{s-1}W_{s-1}))^{-1}$\\
\hline
OMTLVON             &Online multitask perceptron with von-Neumann divergence update for matrix A\\
& $A_s = exp(log A_{s-1} - \eta sym(W^T_{s-1}W_{s-1}))^{-1})$\\
\hline
\end{tabular}
}
\end{center}
\vspace{-0.2cm}
\end{table}

\vspace{-0.2cm}
\subsection{Error rates}
\vspace{-0.2cm}
The averaged results of the error rate and standard deviations from 20 random permutations are reported in Table \ref{error_std}. According to the results, BatchOPT has the lowest error rate on the spam email dataset (average error rate, 11.32\%), and OMTLCOV has the lowest performance on the spam dataset (average error rate, 15.33\%). For the EEG dataset, when classifying whether a worker is resting or active, OMTLCOV has the best accuracy (average error rate, 33.25\%) and CMTL has the worst performance (average error rate, 35.19\%). When classifying whether a worker is relaxed or stressed out, OMTLCOV still has the best accuracy (average error rate, 29.50\%), and BatchOPT gives us the worst classification (average error rate, 35.08\%). Moreover, for spam data, the BatchOPT seems the most stable (lowest standard deviation when permuting data) and for EEG data, OMTLCOV outperforms the others by both accuracy and stability.

\begin{figure*}[!t]
    \begin{minipage}[b]{0.3\textwidth}
        \centering
        \begin{subfigure}
            \centering
            \includegraphics[width=\linewidth]{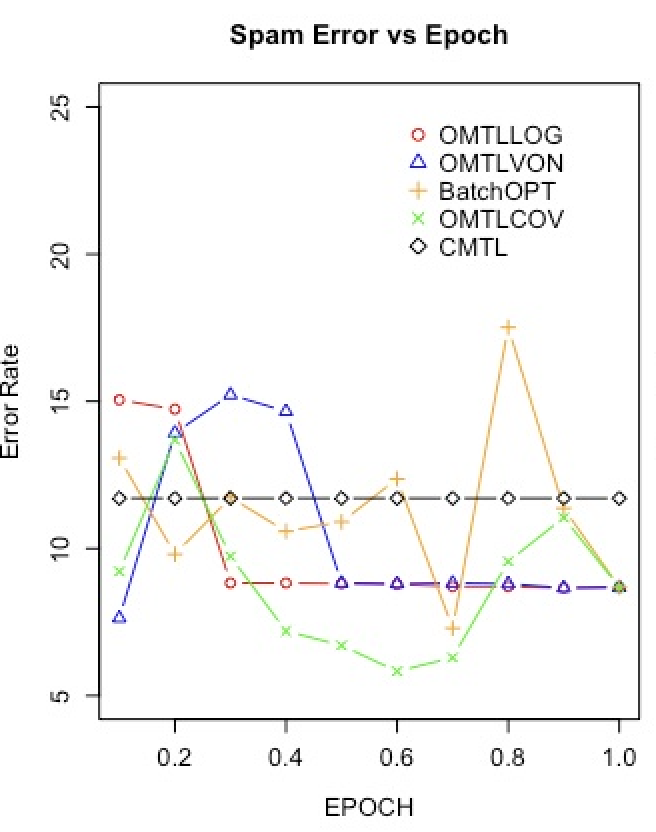}
            \caption{Spam(True/False) error with epoch}
            \label{fig:bfig1}
        \end{subfigure}
    \end{minipage}%
    \vspace{5mm}
    \begin{minipage}[b]{0.3\textwidth}
        \centering
        \begin{subfigure}
            \centering
            \includegraphics[width=\linewidth]{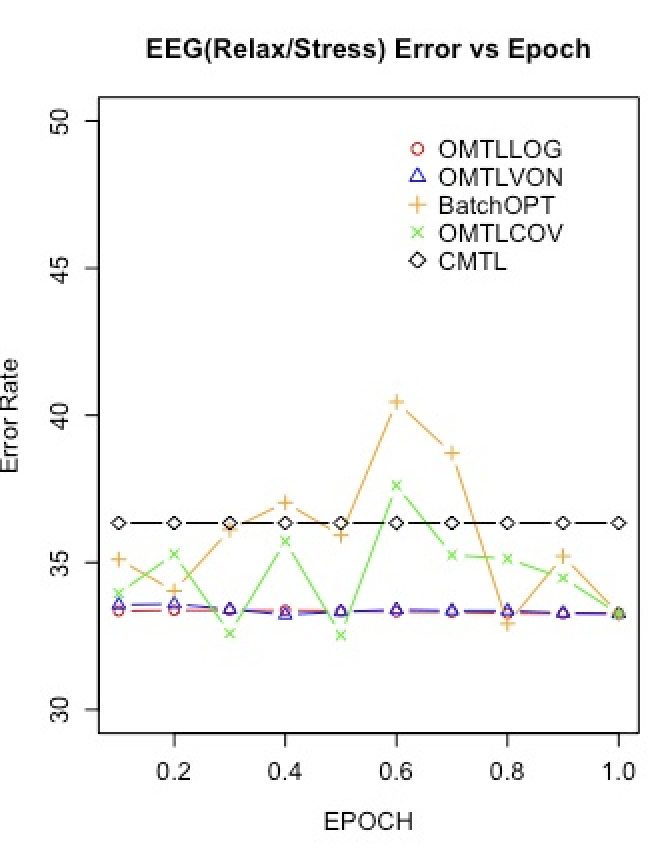}
            \caption{EGG(Relax/Stress) error with epoch}
            \label{fig:bfig2}
        \end{subfigure}
    \end{minipage}%
    \vspace{5mm}
    \begin{minipage}[b]{0.3\textwidth}
        \centering
        \begin{subfigure}
            \centering
            \includegraphics[width=\linewidth]{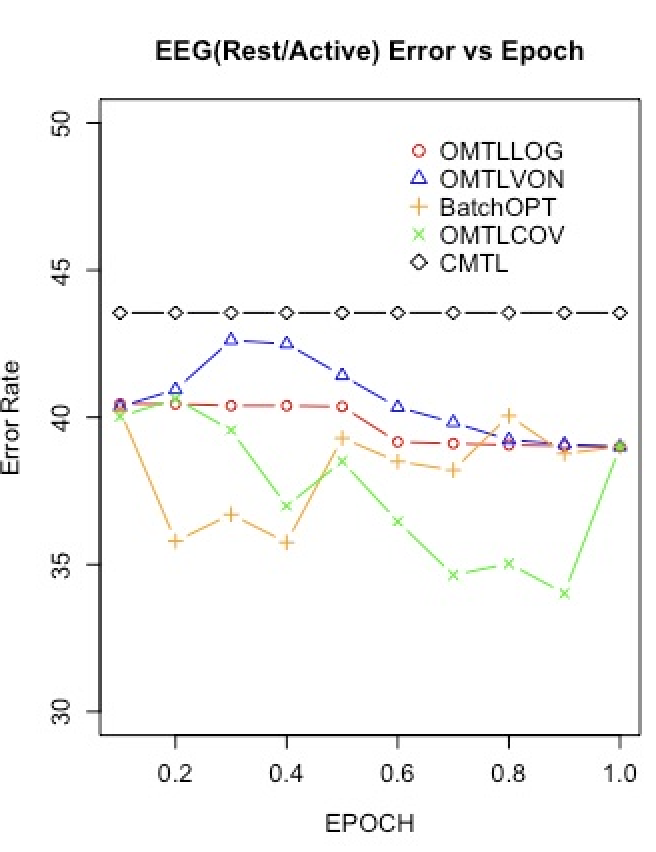}
            \caption{EGG(Rest/Active) error with epoch}
            \label{fig:bfig3}
        \end{subfigure}
    \end{minipage}
\vspace{-1cm}
\end{figure*}

\begin{table}[h]
\caption{Error rates and standard deviations for spam and EEG datasets}
\label{error_std}
\begin{center}
\scalebox{0.87}{
\begin{tabular}{|c|c|c|c|}
 \hline
    \bf Method & \multicolumn{3}{c|}{\bf Error Rate (Standard Deviation) (\%)} \\
    \hline
     & Spam & EEG(Resting/Active) & EEG(Relax/Stressed Out) \\
     & (EPOCH=0.5) & (EPOCH=0.8) & (EPOCH=0.8) \\
    \hline
    CMTL & 12.35 (5.99) & 35.19 (7.54) & 35.05 (11.06) \\
    BatchOPT & \textbf{11.32 (3.46)} & 34.55 (5.83) & 35.08 (6.93) \\
    OMTLCOV & 15.33 (7.29) & \textbf{33.25 (3.04)} & \textbf{29.50 (3.93)} \\
    OMTLLOG & 12.20 (5.66) & 34.66 (4.12) & 33.02 (5.28) \\
    OMTLVON & 12.75 (7.14) & 34.76 (3.87) & 33.01 (5.27) \\
    \hline
\end{tabular}
}
\end{center}
\end{table}

\subsection{Epoch parameter}
As illustrated in Figure \ref{fig:bfig1}, Figure\ref{fig:bfig2} and Figure\ref{fig:bfig3}, adjusting the EPOCH values, which determine the percentage of data seen before updating the relatedness matrix, influences classification errors. OMTLLOG and OMTLVON generally show lower errors with increased EPOCH values. For OMTLCOV, error rates initially decrease then increase, with optimal classification at an EPOCH setting of 0.6. BatchOPT’s prediction accuracy is less affected by changes in EPOCH. An EPOCH range of 0.6-0.7 typically offers the best balance of classification accuracy and computational efficiency, as higher values increase computation time. Notably, the four online learning methods that update the relatedness matrix consistently outperform CMTL, which uses a fixed relatedness matrix, particularly in the EEG datasets.


\subsection{Discussion}
In the spam case, all algorithms demonstrated similar error rates on average. For OMTLLOG and OMTLVON, there were obvious decreasing trends in error rates with increasing EPOCH value, which showed the benefits from adaptive learning and was also consistent with the conclusion of the selected paper: that an increase in Epoch value led to a gradual improvement in prediction accuracy \cite{saha2011online}, although this pattern was not similarly clear for OMTLCOV and BatchOPT. We could also conclude that an EPOCH equal to 0.6-0.7 was a preferable setting in terms of both accuracy and time complexity. For the spam data, all the error rates were lower than what was found in the original paper, which indicated that our dataset differed from theirs and all these algorithms could get similar good accuracy. One possible reason was that the spam data we worked on was balanced, containing 50\% of the spam emails, which might not be same as the dataset used in the paper. In the previous works \cite{saha2011online, liu2024adaptive, wang2024research, liu2024image, liu2024rumor, li2024feature, li2022automated}, they mentioned that different datasets would yield different tipping points and it was reasonable to see different results when testing with different data.

In the EEG case, we observed the OMTLCOV outperformed all other algorithms. OMTLCOV delivered lower and smoother error rates, which validated the simulation results in the paper. However the average error rates were high compared to the spam data. This could be because before the simulation, we needed to extract the features from the brain-test data and the process might potentially bring more skewedness into the experiments, since we only took a limited number of features, which might not give a complete picture of the dataset. Additionally, the sample size was relatively small and the lack of data might be the reason for the unfavorable accuracy. However, due to the capacity of the online learning algorithms, we would expect much better performance by updating the model parameters when getting more data in the future.

It was also possible that the EEG data was not suitable for these learning algorithms. The traditional SVM provided even lower classification errors. It was probably because the EEG data was much more randomized than the spam data and required a better data cleansing process and feature extracting process. 

Moreover, we realized that the learning rate also played a very important role in the online multi-task learning. The updating methods of the relatedness matrix were sensitive to the learning rate to various degrees. Therefore, choosing a proper learning rate would improve the stability and classification accuracy.

\section{Conclusion}

In this paper, we implemented online multi-task learning algorithms and assessed their efficiency across various datasets, including spam and EEG data. We discovered that performance often depended on the dataset’s composition, as indicated by the reference paper \cite{saha2011online}. While algorithms performed well with structured spam data, they struggled with more randomized EEG data, highlighting the need for effective data cleansing and feature extraction, particularly for EEG. We confirmed that updating the relatedness matrix after sufficient data exposure improves prediction accuracy. We also noted the importance of fine-tuning the epoch parameter and other factors like the initial interaction matrix values and learning rate to optimize classifier performance.


\end{document}